\documentclass{article}
\usepackage{spconf,amsmath,graphicx}
\usepackage{amssymb}

\newcommand{\multiline}[1]{\begin{tabular}{@{}c@{}}#1\end{tabular}}
\title{UCorrect: An Unsupervised Framework for Automatic Speech Recognition Error Correction}
\name{\multiline{Jiaxin Guo*\thanks{*Corresponding author: Jiaxin Guo, jiaxinguo1@huawei.com}, Minghan Wang, Xiaosong Qiao, Daimeng Wei, Hengchao Shang, \\ Zongyao Li, Zhengzhe Yu, Yinglu Li, Chang Su, Min Zhang, Shimin Tao, Hao Yang}}
\address{Huawei Translation Services Center, Beijing, China}

\begin{document}

\maketitle

\begin{abstract}
Error correction techniques have been used to refine the output sentences from automatic speech recognition (ASR) models and achieve a lower word error rate (WER). Previous works usually adopt end-to-end models and has strong dependency on Pseudo Paired Data and Original Paired Data. But when only pre-training on Pseudo Paired Data, previous models have negative effect on correction. While fine-tuning on Original Paired Data, the source side data must be transcribed by a well-trained ASR model, which takes a lot of time and not universal. In this paper, we propose UCorrect, an unsupervised Detector-Generator-Selector framework for ASR Error Correction. UCorrect has no dependency on the training data mentioned before. The whole procedure is first to detect whether the character is erroneous, then to generate some candidate characters and finally to select the most confident one to replace the error character. Experiments on the public AISHELL-1 dataset and WenetSpeech dataset show the effectiveness of UCorrect for ASR error correction: 1) it achieves significant WER reduction, achieves 6.83\% even without fine-tuning and 14.29\% after fine-tuning; 2) it outperforms the popular NAR correction models by a large margin with a competitive low latency; and 3) it is an universal method, as it reduces all WERs of the ASR model with different decoding strategies and reduces all WERs of ASR models trained on different scale datasets. 
\end{abstract}
\begin{keywords}
ASR, Error Correction, Unsupervised, WER
\end{keywords}
\section{Introduction}
\label{sec:intro}

In recent years, error correction has been widely adopted to refine ASR output sentences for further WER reduction. Error correction, a typical sequence to sequence task, takes the sentence transcribed by an ASR model\cite{DBLP:conf/interspeech/GulatiQCPZYHWZW20,DBLP:journals/corr/abs-2012-05481,DBLP:journals/corr/abs-2106-05642} as the source inputs and the ground-truth sentence as the target outputs, and aims to correct the errors in the source.

Previous works on ASR error correction usually adopt end-to-end generation models. These models can be grouped into two categories. The first category is the autoregressive (AR) model \cite{DBLP:conf/nips/VaswaniSPUJGKP17}. These models can achieve a large WER reduction but suffer from a low decoding speed. The second category is the non-autoregressive (NAR) model \cite{DBLP:conf/nips/GuWZ19,DBLP:conf/nips/LengTZXLLQLLL21}. The models can speed up the inference process by means of parallel tokens generation on the target side.

End-to-end correction models training depends on alignment paired data. The paired data consist of two parts. The first part is \textbf{Pseudo Paired Data}. It is usually large-scale and used for pre-training. The source side and the target side of Pseudo Paired Data are first constructed from the same crawled text data, and then randomly delete, insert or replace words in the source text. The sencond part is \textbf{Original Paired Data}. It is usually limited and built based on original limited ASR dataset\cite{DBLP:journals/corr/abs-1709-05522,DBLP:conf/icassp/ZhangLGSYXXBCZW22} for correction models fine-tuning. The source side of the Original Paired Data is the transcription of a well-trained ASR model, and the target side is the ground-truth text data. End-to-end correction models training on these two paired data have following difficulties and problems:

\begin{itemize}
    \item \textbf{The error distribution in the Pseudo Paired Data is not consistent with that in the real scene.} Therefore some models only pre-training on pseudo data even result in worse WER than original ASR outputs.
    \item Constructing the Original Paired Data needs to transcribe ASR datasets by a well-trained ASR model first, \textbf{which takes a lot of time.}
    \item Different ASR models have different bias and may generate different transcribed sentences containing different errors. So the correction model is highly correlated with the transcription ASR model, \textbf{which means the correction model is not universal together with the Original Paired Data.}
\end{itemize}

\begin{figure*}[!t]
\centering
\includegraphics[width=0.95\textwidth]{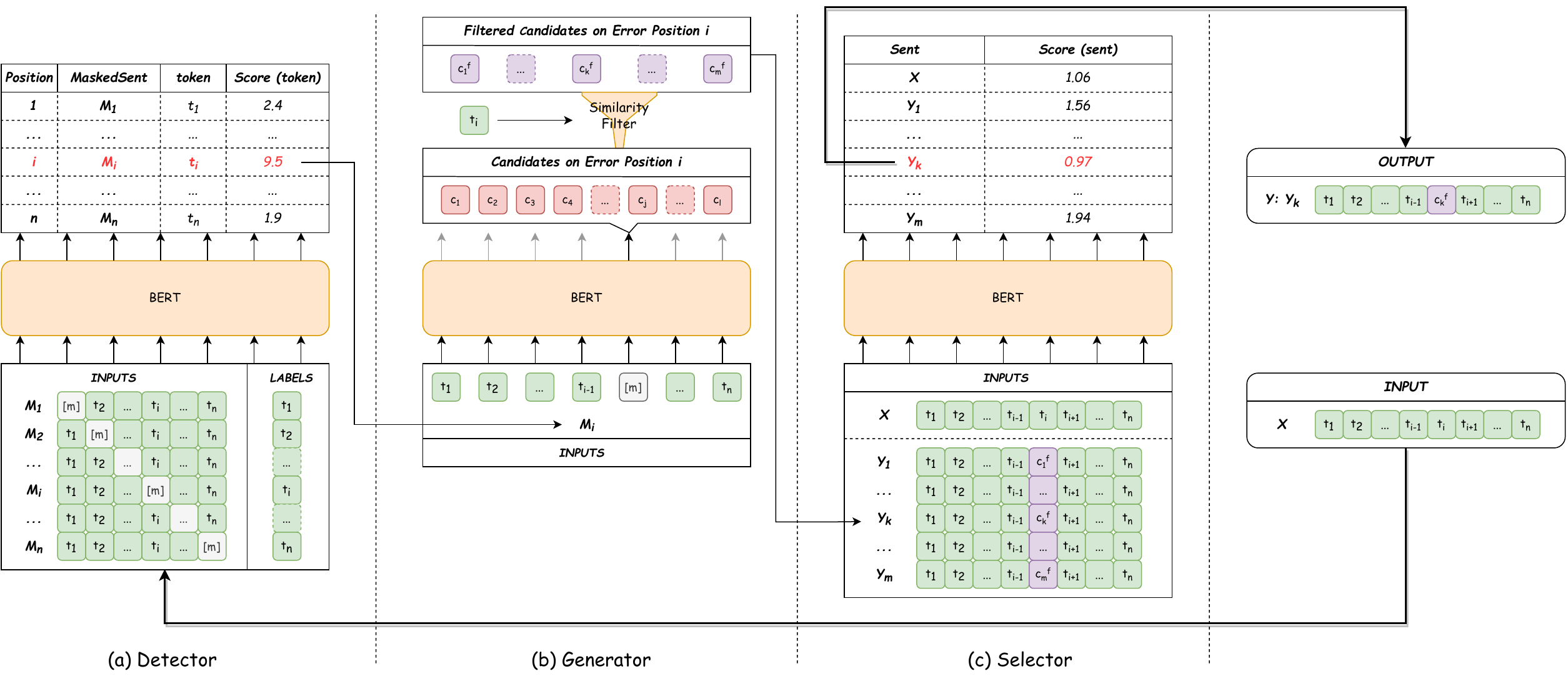} 
\caption{An overview of UCorrect Architecture.}
\label{fig:ucorrect_architecture}
\end{figure*}

\textbf{Moreover, end-to-end models have poor interpretability, and the correction results are uncontrollable}. While producing trustworthy corrections to the erroneous characters, these models correct each character of the sentence regardless of its correctness, \textbf{which might change the correct characters and result in high false alarm rates (FAR).} This issue becomes even more significant when only a small fraction of characters is incorrect. FAR refers to the ratio of error correction numbers among all corrected characters.

In this work, we propose UCorrect, which is a novel BERT-based Detector-Generator-Selector framework for ASR Error Correction. Unlike end-to-end models directly correcting mentioned before, UCorrect first detects whether the character is erroneous, then generates some candidate characters and finally selects the most confident one to replace the error character. Our main contributions are as follows:

\begin{itemize}
    \item \textbf{Our approach is an unsupervised framework, and has no dependency on Pseudo Paired Data. While the end-to-end models only pre-training on Pseudo Paired Data have negative effect on correction, our approach without fine-tuning get positive WER reduction and achieve the lowest WER.}
    \item Our framework is constructed based on pre-training language models (PTM), such as BERT\cite{DBLP:conf/naacl/DevlinCLT19}. We can fine-tune the PTM on the ground-truth text of ASR dataset to achieve further improvements. \textbf{Different from the Original Paired Data, our fine-tuning process only needs the ground-truth data, which saves transcription time.}
    \item \textbf{Our approach significantly improve the correction accuracy with/without fine-tuning, while achieving competitive low latency.}
    \item \textbf{Our approach has better explainability, and can reduce the FAR.} After the errors are detected, the model has a better chance to make the right corrections.
\end{itemize}


\section{Architecture of UCorrect}
\label{sec:architecture}

As shown in Figure \ref{fig:ucorrect_architecture}, UCorrect consists of three parts: a Detector, a Generator and a Selector. Given a sentence $X = (t_1, t_2, ..., t_n)$ consisting of $n$ characters, the Detector predicts the position $i$ of erroneous character and outputs a masked sequence $M_{i} = (t_1, t_2, ..., t_{i_1}, [m], t_{i+1}, ..., t_n)$ with the detected character masked by $[m]$. $[m]$ refers to the special token [MASK] in pre-trained models. The Generator takes the $M_{i}$ as input and generates $m$ candidate characters $\{c_1^f, c_2^f, ..., c_k^f, ..., c_m^f\}$. These $m$ candidate characters are further used to construct $m$ candidate sentences $\{Y_1, Y_2, ..., Y_k, ..., Y_m\}$ by backfilling the candidate character to the masked position $i$. The Selector chooses the best sentence $Y$ as the corrected output among the all $m+1$ possible sentences including $X$.


\subsection{Detector}
\label{ssec:detector}

Given the original input $X$, the Detector outputs a set of scores of each token and predicts the position of erroneous characters according to the token scores.

The Detector takes the following procedure: First, we create a set of masked sentences $\{M_{1}, M_{2}, ..., M_{i}, ..., M_{n}\}$ by replacing each word with the predefined token $[m]$ one at a time, where $M_{i} = (t_1, t_2, ..., t_{i_1}, [m], t_{i+1}, ..., t_n)$. After the creation, we use pre-trained model compute the likelihood of the original character in the masked position as the token score. Finally, we pick the $X_{mi}$ with the highest score. The higher the score, the higher the perplexity of the character, and the higher the error possibility of the original character in the masked position.

The token scoring function can be calculated by:
\begin{align}
    f(t_i) = p(t_i|M_i;\theta)
\end{align}
, where $\theta$ denotes the parameter set of the PTM network.

\subsection{Generator}
\label{ssec:generator}

Given the masked input $M_{i} = (t_1, t_2, ..., t_{i_1}, [m], t_{i+1}, ..., t_n)$, the Generator at last generates $m$ candidates on the masked position.

The Generator has the following three steps: First, the pre-trained model can generates $l$ candidate character sets $\mathcal{C} = \{c_1, c_2, ..., c_j, ..., c_l\}$ at the masked position $i$. Second, according to the similarity to $t_i$, we filter out the most similar $m$ candidates $\mathcal{C}^f = \{c_{1}^f, c_{2}^f, ..., c_{k}^f, ..., c_{m}^f\}$. Third, we replace each $c_k^f$ to $[m]$ of sentence $M_{i}$, and construct $m$ candidate sentences set $\mathcal{Y} = \{Y_1, Y_2, ..., Y_k, ..., Y_m\}$.

The similarity filtering function can be defined as following:
\begin{align}
    f(c_j) = f_{s}(f_{p}(c_j), f_{p}(t_i))
\end{align}
, where $c_j \in \mathcal{C}$, $f_{s}$ represents any phonetic similarity algorithms\cite{DBLP:conf/coling/LiuLCL10} and $f_{p}$ represents any phonetic converter algorithms.

\subsection{Selector}
\label{ssec:selector}

Given the combination sets $\mathcal{Y}\cup\{X\}$, the Selector outputs a set of scores of each candidate sentence and selects the final optimal correction which has the lowest score. The lower the sentence score, the higher the sentence probability.

The sentence scoring method is to calculate the average of token scores of each character in one sentence. Using original text $X = (t_1, t_2, ..., t_n)$ as example, the formulation is as following:
\begin{align}
    f(X) = \frac {\sum_{i=1}^n p(t_i|M_i;\theta)} {n}
\end{align}
, where $\theta$ denotes the parameter set of the PTM network, and $M_{i}$ represents the masked sentence of $X$ on position $i$.

\section{Experimental Setup}
\label{sec:experiments}

\subsection{Datasets}
\label{ssec:Datasets}

We conduct experiments on two public datasets, one is a simple research-oriented dataset AISHELL-1\cite{DBLP:journals/corr/abs-1709-05522}, and the other is a complex production-oriented dataset WenetSpeech\cite{DBLP:conf/icassp/ZhangLGSYXXBCZW22}.

\textbf{AISHELL-1} The AISHELL-1\cite{DBLP:journals/corr/abs-1709-05522} dataset is a Mandarin speech corpus with 178 hours training data, 10 hours validation data and 5 hours test data.

\textbf{WenetSpeech} The WenetSpeech\cite{DBLP:conf/icassp/ZhangLGSYXXBCZW22} dataset is a 10000+ hours multi-domain transcribed Mandarin Speech Corpus. This dataset have 10 categories according to speaking styles and spoken scenarios. The validation data is about 20 hours and the test data is about 23 hours.

\subsection{ASR models}
\label{ssec:asr_models}

To verify the universality of our approach, we conduct our experiments on two popular ASR models. One is Conformer, and the other is U2++ supporting multiple decoding strategies.

\textbf{Conformer} Follow the previous work, we use the ESPnet\cite{DBLP:conf/interspeech/WatanabeHKHNUSH18} toolkit to train a Conformer\cite{DBLP:conf/interspeech/GulatiQCPZYHWZW20} model on AISHELL-1 dataset. Several techniques such as SpecAugment and speed perturbation are utilized to improve the performance of this ASR model.


\textbf{U2++} We use the Wenet\cite{DBLP:conf/interspeech/ZhangWPSY00YP022} toolkit to train multiple U2++\cite{DBLP:journals/corr/abs-2106-05642} Conformer models on different datasets. U2++ model is an enhanced version of U2\cite{DBLP:journals/corr/abs-2012-05481} model, which is a novel two-pass approach to unify streaming and non-streaming end-to-end speech recognition in a single model. U2++ adopts the hybrid CTC/attention architecture, and uses a dynamic chunk-based attention strategy to allow arbitrary right context length. U2++ supports four decoding strategies, which are "ctc greedy search", "ctc prefix beam search", "attention decoder" and "attention rescoring".

\subsection{Related Error Correction Systems}
\label{ssec:configurations_and_baseline}

We describe the several related systems for comparison: two non-autoregressive (NAR) models, LevT\cite{DBLP:conf/nips/GuWZ19} and FastCorrect\cite{DBLP:conf/nips/LengTZXLLQLLL21}, and a Transformer\cite{DBLP:conf/nips/VaswaniSPUJGKP17} based autoregressive (AR) model. All these systems are all end-to-end models. The model configurations follow the previous work \cite{DBLP:conf/nips/LengTZXLLQLLL21}.

\subsection{Fine-tuning}
\label{ssec:fine_tuning}

As is known to all, the pre-training procedure of BERT\cite{DBLP:conf/naacl/DevlinCLT19} includes two unsupervised tasks. One is Masked LM task, the other is Next Sentence Prediction task. During our fine-tuning procedure, the ground-truth text of ASR dataset is used to train on the Masked LM task with masked ratio of 20\%.

\begin{table}[htbp]
\centering
\begin{tabular}{l|c|cc}
\hline
\hline
\textbf{Model} & \textbf{FAR} & \multicolumn{2}{c}{\textbf{Test}} \\
  & & \textbf{wer} & \textbf{werr} \\
\hline
No correction & - & 4.83 & - \\
\hline
\hline
AR model* & - & 5.28 & -9.32  \\
FastCorrect* & - & 5.19 & -7.45 \\
\hline
UCorrect* & \textbf{44.1} & \textbf{4.50} & \textbf{6.83} \\
\hline
\hline
AR model      & 46.0 &  \textbf{4.08} & \textbf{15.53}  \\
LevT(MIter=1) & 61.6 &  4.73 &  2.07  \\
LevT(MIter=3) & -    &  4.74 &  1.86  \\
FastCorrect   & 49.9 &  4.16 & 13.87  \\
\hline
UCorrect & \textbf{41.6} & 4.14 & 14.29 \\
\hline
\end{tabular}
\caption{Evaluation of accuracy on AISHELL-1 dataset generated by a Conformer ASR model under different correction models. (\textbf{Notes:} * indicate the results are obtained without fine-tuning.)}
\label{tab:eval_accuracy}
\end{table}

\section{Results and Discussion}
\label{sec:results}

\subsection{Comparison of UCorrect and Related Systems}
\label{ssec:eval_aishell1_conformer}

We first report the accuracy and latency of UCorrect comparing to baseline error correction models on AISHELL-1 dataset generated by a Conformer ASR model in Table ~\ref{tab:eval_accuracy} and Table ~\ref{tab:eval_latency}.

We have several observations: 1) Without fine-tuning, our approach can reduce the WER (measured by WERR) of ASR model by 4.50\% on the test set of AISHELL-1. Otherwise end-to-end models result in worse WER than original ASR outputs. 2) After fine-tuning, all systems achieve WER reduction. Our approach outperforms all NAR models. 3) Our approach reduces the FAR significantly and achieves the lowest FAR. The FAR of UCorrect before fine-tuning is even lower than that of end-to-end models after fine-tuning.

\begin{table}[htbp]
\centering
\begin{tabular}{l|c|cc}
\hline
\hline
\textbf{Model} & \textbf{Test} & \multicolumn{2}{c}{\textbf{Latency}} \\
  & \textbf{wer} & \textbf{ms/sent} & \textbf{speedup} \\
\hline
AR model      &  4.08 & 149.5 & 1$\times$  \\
LevT(MIter=1) &  4.73 &  54.0 & 2.77$\times$  \\
LevT(MIter=3) &  4.74 &  60.5 & 2.47$\times$  \\
FastCorrect   &  4.16 & \textbf{21.2} & 7.05$\times$  \\
\hline
UCorrect & 4.14 & 35.12 & 4.26$\times$ \\
\hline
\end{tabular}
\caption{Evaluation of latency on AISHELL-1 dataset generated by the Conformer ASR model under different correction models.}
\label{tab:eval_latency}
\end{table}

Table ~\ref{tab:eval_latency} shows that our proposed UCorrect  speeds up the inference of the AR model by 4+ times on the dataset AISHELL-1, which is a competitive low latency comparing to other systems.

\subsection{Evaluation under ASR outputs with different decoding strategies}
\label{ssec:eval_aishell1_u2plus}

We adopt our proposed UCorrect to different ASR outputs with different decoding strategies. We evaluate the accuracy on AISHELL-1 dataset generated by a U2++ ASR model and the results are shown in Table ~\ref{tab:eval_aishell1_u2plus}.

\begin{table}[htbp]
\centering
\begin{tabular}{@{}l|c|cc@{}}
\hline
\hline
\textbf{Decoding Strategy} & \textbf{No Correction} & \multicolumn{2}{c}{\textbf{UCorrect*}} \\
       & \textbf{wer}   & \textbf{wer}       & \textbf{werr}   \\
\hline
ctc greedy search       & 4.94 & \textbf{4.58} & 7.29 \\
ctc prefix beam search  & 4.94 & \textbf{4.59} & 7.09 \\
attention decoder       & 5.21 & \textbf{4.96} & 4.80 \\
attention rescoring     & 4.62 & \textbf{4.31} & 6.71 \\
\hline
\end{tabular}
\caption{Evaluation of accuracy on AISHELL-1 dataset generated by the U2++ ASR model with different decoding strategies. (\textbf{Notes:} * indicate the results are obtained without fine-tuning.)}
\label{tab:eval_aishell1_u2plus}
\end{table}

We have several observations: 1) Our proposed UCorrect reduce all WERs of the ASR model with different decoding strategies and achieves 4-8\% WERR, which proves our method is universal. 2) Our proposed UCorrect reduces the WER of "ctc greedy search" results at most and reduces the WER of "attention decoder" results at least.

\subsection{Evaluation under different ASR datasets}
\label{ssec:eval_wenet_u2plus}

We adopt our proposed UCorrect to different ASR datasets. We evaluate the accuracy on AISHELL-1 dataset and WenetSpeech dataset generated by a U2++ ASR model with "ctc greedy search" decoding strategy and the results are shown in Table ~\ref{tab:eval_wenet_u2plus}.

\begin{table}[htbp]
\centering
\begin{tabular}{@{}l|cc|cc@{}}
\hline
\hline
\textbf{Model} & \multicolumn{2}{c}{\textbf{AISHELL-1}} & \multicolumn{2}{c}{\textbf{WenetSpeech}} \\
       & \textbf{wer} & \textbf{werr}   & \textbf{wer} & \textbf{werr}   \\
\hline
No correction & 4.94 & - & 9.68 & - \\
\hline
UCorrect* & \textbf{4.58} & 7.29 & \textbf{9.41} & 2.79 \\
UCorrect & \textbf{4.20} & 14.98 & \textbf{9.24} & 4.55 \\
\hline
\multicolumn{4}{c}{} \\
\end{tabular}
\caption{Evaluation of accuracy on AISHELL-1 dataset and WenetSpeech dataset generated by the U2++ ASR model. (\textbf{Notes:} * indicate the results are obtained without fine-tuning.)}
\label{tab:eval_wenet_u2plus}
\end{table}

We have several observations: 1) Our proposed UCorrect reduce all WERs of ASR models trained on different datasets, which proves our method is universal. 2) Our proposed UCorrect achieves 14.98\% WERR on AISHELL-1 dataset and 4.55\% on WenetSpeech dataset, as WenetSpeech is a large-scale dataset and more complex.


\section{Conclusion}
In this paper, we propose UCorrect, an unsupervised framework for ASR Error Correction. UCorrect consists of three parts: a Detector, a Generator and a Selector. The Detector detects the erroneous position in the sentence; the Generator generates candidate characters on the detected position; and the Selector selects the most confident one as output. Experiments on the various ASR datasets under various decoding strategies show the effectiveness of UCorrect. Without fine-tuning, our proposed UCorrect achieves the state-of-the-art performance. After fine-tuning, our proposed UCorrect outperforms the popular NAR correction models by a large margin with a competitive low latency.

\bibliographystyle{IEEEbib}
\bibliography{main}

\begin{thebibliography}{10}

\bibitem{DBLP:conf/interspeech/GulatiQCPZYHWZW20}
Anmol Gulati, James Qin, Chung{-}Cheng Chiu, Niki Parmar, Yu~Zhang, Jiahui Yu,
  Wei Han, Shibo Wang, Zhengdong Zhang, Yonghui Wu, and Ruoming Pang,
\newblock ``Conformer: Convolution-augmented transformer for speech
  recognition,''
\newblock in {\em Interspeech 2020, 21st Annual Conference of the International
  Speech Communication Association, Virtual Event, Shanghai, China, 25-29
  October 2020}, Helen Meng, Bo~Xu, and Thomas~Fang Zheng, Eds. 2020, pp.
  5036--5040, {ISCA}.

\bibitem{DBLP:journals/corr/abs-2012-05481}
Binbin Zhang, Di~Wu, Zhuoyuan Yao, Xiong Wang, Fan Yu, Chao Yang, Liyong Guo,
  Yaguang Hu, Lei Xie, and Xin Lei,
\newblock ``Unified streaming and non-streaming two-pass end-to-end model for
  speech recognition,''
\newblock {\em CoRR}, vol. abs/2012.05481, 2020.

\bibitem{DBLP:journals/corr/abs-2106-05642}
Di~Wu, Binbin Zhang, Chao Yang, Zhendong Peng, Wenjing Xia, Xiaoyu Chen, and
  Xin Lei,
\newblock ``{U2++:} unified two-pass bidirectional end-to-end model for speech
  recognition,''
\newblock {\em CoRR}, vol. abs/2106.05642, 2021.

\bibitem{DBLP:conf/nips/VaswaniSPUJGKP17}
Ashish Vaswani, Noam Shazeer, Niki Parmar, Jakob Uszkoreit, Llion Jones,
  Aidan~N. Gomez, Lukasz Kaiser, and Illia Polosukhin,
\newblock ``Attention is all you need,''
\newblock in {\em Advances in Neural Information Processing Systems 30: Annual
  Conference on Neural Information Processing Systems 2017, December 4-9, 2017,
  Long Beach, CA, {USA}}, Isabelle Guyon, Ulrike von Luxburg, Samy Bengio,
  Hanna~M. Wallach, Rob Fergus, S.~V.~N. Vishwanathan, and Roman Garnett, Eds.,
  2017, pp. 5998--6008.

\bibitem{DBLP:conf/nips/GuWZ19}
Jiatao Gu, Changhan Wang, and Junbo Zhao,
\newblock ``Levenshtein transformer,''
\newblock in {\em Advances in Neural Information Processing Systems 32: Annual
  Conference on Neural Information Processing Systems 2019, NeurIPS 2019,
  December 8-14, 2019, Vancouver, BC, Canada}, Hanna~M. Wallach, Hugo
  Larochelle, Alina Beygelzimer, Florence d'Alch{\'{e}}{-}Buc, Emily~B. Fox,
  and Roman Garnett, Eds., 2019, pp. 11179--11189.

\bibitem{DBLP:conf/nips/LengTZXLLQLLL21}
Yichong Leng, Xu~Tan, Linchen Zhu, Jin Xu, Renqian Luo, Linquan Liu, Tao Qin,
  Xiangyang Li, Edward Lin, and Tie{-}Yan Liu,
\newblock ``Fastcorrect: Fast error correction with edit alignment for
  automatic speech recognition,''
\newblock in {\em Advances in Neural Information Processing Systems 34: Annual
  Conference on Neural Information Processing Systems 2021, NeurIPS 2021,
  December 6-14, 2021, virtual}, Marc'Aurelio Ranzato, Alina Beygelzimer,
  Yann~N. Dauphin, Percy Liang, and Jennifer~Wortman Vaughan, Eds., 2021, pp.
  21708--21719.

\bibitem{DBLP:journals/corr/abs-1709-05522}
Hui Bu, Jiayu Du, Xingyu Na, Bengu Wu, and Hao Zheng,
\newblock ``{AISHELL-1:} an open-source mandarin speech corpus and {A} speech
  recognition baseline,''
\newblock {\em CoRR}, vol. abs/1709.05522, 2017.

\bibitem{DBLP:conf/icassp/ZhangLGSYXXBCZW22}
Binbin Zhang, Hang Lv, Pengcheng Guo, Qijie Shao, Chao Yang, Lei Xie, Xin Xu,
  Hui Bu, Xiaoyu Chen, Chenchen Zeng, Di~Wu, and Zhendong Peng,
\newblock ``{WENETSPEECH:} {A} 10000+ hours multi-domain mandarin corpus for
  speech recognition,''
\newblock in {\em {IEEE} International Conference on Acoustics, Speech and
  Signal Processing, {ICASSP} 2022, Virtual and Singapore, 23-27 May 2022}.
  2022, pp. 6182--6186, {IEEE}.

\bibitem{DBLP:conf/naacl/DevlinCLT19}
Jacob Devlin, Ming{-}Wei Chang, Kenton Lee, and Kristina Toutanova,
\newblock ``{BERT:} pre-training of deep bidirectional transformers for
  language understanding,''
\newblock in {\em Proceedings of the 2019 Conference of the North American
  Chapter of the Association for Computational Linguistics: Human Language
  Technologies, {NAACL-HLT} 2019, Minneapolis, MN, USA, June 2-7, 2019, Volume
  1 (Long and Short Papers)}, Jill Burstein, Christy Doran, and Thamar Solorio,
  Eds. 2019, pp. 4171--4186, Association for Computational Linguistics.

\bibitem{DBLP:conf/coling/LiuLCL10}
Chao{-}Lin Liu, Min{-}Hua Lai, Yi{-}Hsuan Chuang, and Chia{-}Ying Lee,
\newblock ``Visually and phonologically similar characters in incorrect
  simplified chinese words,''
\newblock in {\em {COLING} 2010, 23rd International Conference on Computational
  Linguistics, Posters Volume, 23-27 August 2010, Beijing, China}, Chu{-}Ren
  Huang and Dan Jurafsky, Eds. 2010, pp. 739--747, Chinese Information
  Processing Society of China.

\bibitem{DBLP:conf/interspeech/WatanabeHKHNUSH18}
Shinji Watanabe, Takaaki Hori, Shigeki Karita, Tomoki Hayashi, Jiro Nishitoba,
  Yuya Unno, Nelson Enrique~Yalta Soplin, Jahn Heymann, Matthew Wiesner, Nanxin
  Chen, Adithya Renduchintala, and Tsubasa Ochiai,
\newblock ``Espnet: End-to-end speech processing toolkit,''
\newblock in {\em Interspeech 2018, 19th Annual Conference of the International
  Speech Communication Association, Hyderabad, India, 2-6 September 2018},
  B.~Yegnanarayana, Ed. 2018, pp. 2207--2211, {ISCA}.

\bibitem{DBLP:conf/interspeech/ZhangWPSY00YP022}
Binbin Zhang, Di~Wu, Zhendong Peng, Xingchen Song, Zhuoyuan Yao, Hang Lv, Lei
  Xie, Chao Yang, Fuping Pan, and Jianwei Niu,
\newblock ``Wenet 2.0: More productive end-to-end speech recognition toolkit,''
\newblock in {\em Interspeech 2022, 23rd Annual Conference of the International
  Speech Communication Association, Incheon, Korea, 18-22 September 2022},
  Hanseok Ko and John H.~L. Hansen, Eds. 2022, pp. 1661--1665, {ISCA}.

\end{thebibliography}

\end{document}